\documentclass[conference]{IEEEtran}
\IEEEoverridecommandlockouts
\usepackage{cite}
\usepackage{amsmath,amssymb,amsfonts}
\usepackage{multirow}
\usepackage{graphicx}
\usepackage{textcomp}
\usepackage{algorithm}
\usepackage{algpseudocode}
\usepackage{xcolor}
\usepackage[caption=false]{subfig}

\def\BibTeX{{\rm B\kern-.05em{\sc i\kern-.025em b}\kern-.08em
    T\kern-.1667em\lower.7ex\hbox{E}\kern-.125emX}}
\begin{document}

\title{A Multi-Objective approach to the Electric Vehicle Routing Problem}

\author{
\IEEEauthorblockN{ Kousik Rajesh}
\IEEEauthorblockA{\textit{Deptartment of Computer Science and Engineering} \\
\textit{Indian Institute of Technology Guwahati}} \and
\IEEEauthorblockN{Eklavya Jain}
\IEEEauthorblockA{\textit{Department of Mathematics} \\
\textit{Indian Institute of Technology Guwahati}}
\and
\IEEEauthorblockN{Prakash Kotecha}
\IEEEauthorblockA{\textit{Department of Chemical Engineering} \\
\textit{Indian Institute of Technology Guwahati}\\}

}

\maketitle

\begin{abstract}
The electric vehicle routing problem (EVRP) has garnered great interest from researchers and industrialists in an attempt to move from fuel-based vehicles to healthier and more efficient electric vehicles (EVs). While it seems that the EVRP should not be much different from traditional vehicle routing problems (VRPs), challenges like limited cruising time, long charging times, and limited availability of charging facilities for electric vehicles makes all the difference. Previous works target logistics and delivery-related solutions wherein a homogeneous fleet of commercial EVs have to return to the initial point after making multiple stops. On the opposing front, we solve a personal electric vehicle routing problem and provide an optimal route for a single vehicle in a long origin-destination (OD) trip. We perform multi-objective optimization - minimizing the total trip time and the cumulative cost of charging. In addition, we incorporate external and real-life elements like traffic at charging stations, detour distances for reaching a charging station, and variable costs of electricity at different charging stations into the problem formulation. In particular, we define a multi-objective mixed integer non-linear programming (MINLP) problem and obtain a feasible solution using the $\epsilon$-constraint algorithm. We further implement meta-heuristic techniques such as Genetic Algorithm (GA) and Particle Swarm Optimization (PSO) to obtain the most optimal route and hence, the objective values. The experiment is carried out for multiple self-generated data instances and the results are thereby compared. 
\end{abstract}


\section{Introduction}
There are over 100 electric car manufactures in the world, over 500 electric vehicle-related startups in India alone, and a lot more around the world. Electric Vehicles have the potential to revolutionize, not only the automobile industry, but the entire world owing to its green effects on the environment, cheaper running costs, and negligible noise pollution. Although the manufacturing of EVs itself takes a lot of energy, they are still a greener option considering the significantly reduced emissions of carbon dioxide during its lifetime. Many logistics companies like Mahindra Logistics Limited (MLL), Deutsche Post DHL Group, and even Amazon have already shifted to electric vehicles. 

The traditional EVRP finds an optimal route for the scenario where a fleet of electric vehicles have to visit multiple delivery locations, recharge at a few stops, and ultimately return to the depot node. Due to the diversity of the problem, there exist innumerable variations to the problem. Although, a few basic assumptions that are common to the work in \cite{LIN2016508}, \cite{felipe2014heuristic}, and \cite{paz2018multi} are discussed in \cite{KUCUKOGLU2021107650}. Kucukoglu et al. \cite{KUCUKOGLU2021107650} categorize the domain based on the possible objective functions, potential constraints, and various charging policies. 

The EVRP can also be viewed as an extension to the green vehicle routing problem proposed by Erdogan and Miller-Hooks \cite{ERDOGAN2012100} in which they consider alternative fuel vehicles (including EVs) that have limited travel range and must recharge during the route. They minimize the distance traveled by vehicles in their research. We adopt their underlying assumption of requiring at least one recharge of the vehicle while traveling from the source to the destination. However, we consider separate charging times at different charging stations depending upon the technology used at a station, and the possibility of partial recharging of vehicles in order to minimize cost of refueling throughout the trip. 

Following Erdogan, and Miller-Hooks, Felipe et al. \cite{felipe2014heuristic} and Ding et al. \cite{ding2015conflict} proposed an EVRP with multiple charging technologies and partial recharging. They consider the energy consumption to be a function of distance traveled, and optimize the total distance traveled for a homogeneous fleet of electric vehicles. Likewise, we assume the energy consumption to be a linear function of distance traveled between any two nodes in the underlying topology graph.

Furthermore, electric vehicles often charge in a non-linear manner and discharge depending upon the geography and the load on the vehicle. Kancharla and Ramadurai \cite{kancharla2020electric} adopt similar constraints while formulating their version of EVRP. Unlike their formulation, where they assume an unlimited fleet of homogeneous EVs and homogeneous charging station in terms of cost of electricity and rate of charging, we consider a single electric vehicle and heterogeneous costs and speeds of recharging. 

In summary, the traditional EVRP is dedicated for scheduling a fleet of electric vehicles to numerous charging stations with varying characteristics. The underlying assumptions considered in previous works associated with EVRP don't fit into the regime of personal electric vehicle routing and thus we have addressed this problem. Liu et al. \cite{liu2014joint} consider a joint charging and routing optimization problem for electric vehicles. They consider a real-time pricing model at charging stations to motivate load balancing in the central distribution grid. They provide a deterministic algorithmic solution to the long origin-destination problem, but do not consider the possibility of non-linear charging rates, heterogeneity of chargers, detour distances associated with each charging stations, and delays due to external traffic. 

In existing literature, EV \textit{navigation} has been studied in multiple works. For instance, \cite{sachenbacher2011efficient} discusses an energy-optimal routing problem with the consideration of battery limitations, dynamic edge costs, and recuperation, etc.. \cite{de2013intention} describes a novel technique to combine EV drivers' intentions and historical information of charging stations to predict queues at charging stations. These predictions along with direct information from charging stations can help in estimating the expected waiting time for each charging station that we consider as a parameter in our problem formulation. However, the solution to that problem only focuses on providing a path with minimum delay, which may or may not be monetarily feasible for everyone. 

In our study to be presented, we propose a general, yet modified EVRP formulation for personal EVs, that considers (1) different starting and end points, (2) variable cost of electricity at different charging stations, (3) multiple speeds of charging at each station, (4) multi-objective optimization problem - minimizing time and cost of travel, (5) detour distance to reach the charging station, (6) external traffic in the form of expected waiting times at each charging station, and (7) partial recharging of the electric vehicle. On the other hand, we assume a directed acyclic graph as the underlying network topology. However, that can be readily extended to function on a general graph topology by creating multiple independent paths for each possible pair of nodes. 

% Our contributions are as follows:
% \begin{enumerate}
%     \item We formulate a modified EVRP for personal vehicles while considering various external factors like variable pricing, traffic delays, and detour distances required to reach charging stations. 
%     \item We implement single objective MINLPs on GAMS using various solvers and compare them over 4 self-generated data instances. 
%     \item We extend the model to multi-objective optimization problem and implement the $\epsilon$-constraint method to obtain a set of non-dominated solutions to the MINLP. We further drew the Pareto front. 
%     \item Finally, we define a solution structure for meta-heuristic techniques, and implement GA and PSO for solving our formulation. We incorporate a penalty for each violated constraint, and plot the convergence curve for each objective function. 
%     \item We compare the solutions of the $\epsilon$-constraint method and the meta-heuristic technique.
% \end{enumerate}
The organization of this paper is as follows. Section \ref{sec:problemdef} explains the problem in depth and introduces the mathematical formulation. In the following section, Section \ref{sec:mathform}, we formulate our problem as a mixed integer non-linear programming (MINLP) problem. The self-generated datasets used to test our formulation are described in Section \ref{sec:dataset}. We solve the MINLP in the GAMS software and extend single objective optimization solutions to multi-objective optimization Pareto fronts using $\epsilon$-constraint method in Section \ref{sec:mathprog}. We compile our results of mathematical programming in Section \ref{sec:results}. Section \ref{sec:metaheuristic} discusses the meta-heuristic technique employed to solve the problem formulation and corresponding results. We conclude with our findings and observations in Section \ref{sec:conclusions}. 
% Finally, we discuss the potential future work that can be done to improve upon the presented work in Section \ref{sec:future}.

\section{Problem Definition}
\label{sec:problemdef}
Consider a directed acyclic graph (DAG), $G = (V, E)$, where $V$ is the set of charging stations, and $E$ is the set of edges. An edge between nodes $i$ and $j$ has an associated weight which signifies the distance from node $i$ to node $j$. This distance between the node $i \in V$, and $j \in V$ is denoted by $d_{ij}$. An electric vehicle with battery capacity $C$ and mileage $\gamma$, starts from a source node $S \not\in V$ and has to travel to a destination node $D \not\in V$. The distance of each $i \in V$ from the source node $S$, and the destination node $D$ is given by $d_{Si}$, and $d_{iD}$, respectively. Each charging station (or node in $G$) has variable cost of recharging based on local electricity prices, denoted by $\psi_i,\ \forall i \in V$. A charging station $i \in V$ is also associated with an expected waiting time, $w_i$, and a detour distance $\delta_i$ in appropriate units. In addition to this, a charging station can be of three types - $L_1$, $L_2$, or $L_3$, depending upon the speed of charging offered at each station. 

Concisely, the problem is to find an optimal route between the source node $S$ and the destination node $D$ ensuring that the vehicle stops at charging stations $\{i, j, \dots\} \subseteq V$ such that the cumulative time taken for the trip including charging times and waiting times is minimized as well as the total cost of charging is minimized. The vehicle can either be fully recharged, partially recharged, or not recharged at all at a node, i.e., a charging station can just be a transit node. 

\section{Mathematical Formulation}
\label{sec:mathform}

\begin{table}
\label{tab:decision_variables}
\centering
\caption{Decision Variables} 
\begin{tabular}{p{0.1\linewidth}p{0.55\linewidth}p{0.2\linewidth}}
\hline
Variable & Definition & Type\\
\hline
% path attributes
$z_i$ & Indicates if the vehicle visited charging station $i$ & Binary \\
$y_i$ & Amount of charge gained at charging station $i$  & Continuous\\
$q_i$ & Vehicle state-of-charge when leaving node $i$ & Continuous\\
$x_{ij}$ & Indicates if the vehicle traveled from charging station $i$ to charging station $j$ & Binary\\
$w^S_i$ & Indicates if vehicle visited charging station $i$ from starting node $S$ & Binary \\
$w^D_i$ & Indicates if vehicle reached the destination node $D$ from the charging station $i$ & Binary \\
$\beta$ & Denotes the final charge left in the vehicle at the destination & Continuous\\

\hline
\end{tabular}
\end{table}

\begin{table}
\label{tab:attributes}
\centering
\caption{Attributes} 
\begin{tabular}{p{0.1\linewidth}p{0.83\linewidth}}
\hline
Attribute & Description\\
\hline
%  vehicle attributes
$v$ & Average speed of the vehicle \\
$\gamma$ & Mileage of the vehicle (distance per unit charge) \\
$C$ & Battery capacity of the vehicle \\
$\alpha$ & Denotes the initial charge in the vehicle at the Source \\
% charging station attributes
$L_i$ & Level of charger at charging station $i$  \\
$\psi_i$ & Cost of recharging at charging station $i$  \\
$w_i$ & Waiting time at charging station $i$   \\
$\delta_i$ & Detour distance required for charging station $i$  \\
$d_{ij}$ & Distance between charging station $i$ and charging station $j$  \\
$d_{Si}$ & Distance of the charging station $i$ from the source node $S$\\
$d_{iD}$ & Distance of the destination node $D$ from the charging station $i$ \\
\hline
\end{tabular}
\end{table}

\subsection{Variables}
We formulate an MINLP to model the modified EVRP. We define various decision variables as defined in Table I.  Some other important attributes have been described in Table II.

\subsection{Objective Functions}
The first objective of minimizing travel time is given by 
\begin{multline}
   min\ \ \sum_{i \in V} \frac{w^S_i  d_{Si}}{v} + \sum_{i \in V} \sum_{j \in V} \frac{x_{ij}d_{ij}}{v} + \sum_{i \in V} \frac{w^D_i  d_{iD}}{v} \\ + \sum_{i \in V} \frac{z_i \delta_i}{v} + \sum_{i \in V} z_i w_i + \sum_{i \in V} \frac{z_i y_i C}{L_i}.
   \label{eqn:obj_time}
\end{multline}

The first three terms of Eqn. \ref{eqn:obj_time} denote the time taken to reach the destination node $D$ from the source node $S$ when routed through the given network of charging stations $G = (V, E)$. The last three terms are associated with the charging process. $w^S_i$ indicates whether the vehicle visited charging station $i \in V$ from the source node $S$. So the first term signifies the time taken to travel from the source node $S$ to the first node in $V$. The second term calculates the cumulative time taken to reach the penultimate node in $G$. We use the binary variable $x_{ij}$ to indicate whether the edge $ij \in E$ was traveled by the vehicle or not. The third term uses $w^D_i$ to give the time taken to reach the ultimate destination node $D$ from the penultimate node. Another decision variable $z_i$ is used to indicate whether the vehicle visited the charging station $i \in V$ or not. The continuous variable $y_i$ denotes the amount of charge added to the vehicle at charging station $i \in V$. The fourth term corresponds to the time taken for the detours made at all the charging stations where the vehicle recharged its battery. The fifth term is the waiting time associated with the charging stations where the vehicle was recharged. Finally, the sixth term is the time taken to recharge the vehicle by amount $y_iC$ with speed of charging as $L_i$. \\

The second objective of minimizing cost is given by
\begin{equation}
    min\ \ \sum_{i \in V} z_i y_i C \psi_i.
    \label{eqn:obj_cost}
\end{equation}

The only cost is corresponding to the charging of the vehicle at various charging stations. Each charging station $i \in V$ is associated with a different electricity rate $\psi_i$. We calculate the total cost by considering the charging cost of the stations where the vehicle stopped and recharged its battery.  

\subsection{Constraints} 

\textbf{Source out-degree constraint}: This constraint enforces that the source node should be departed from exactly one node.
\begin{equation}
    \sum_{i \in V} w^S_i = 1 
\end{equation}

\textbf{Destination in-degree constraint}: This constraint enforces that the destination node should be arrived on from exactly one node.
\begin{equation}
    \sum_{i \in V} w^D_i = 1 
\end{equation}

\textbf{In-degree constraint}: The path picked should be a valid path i.e Each node should be arrived on from no more than one other node. This is enforced by summing up $x_{ij}$ for all $i \in V$ and constraining it to be $\leq 1$. 
\begin{equation}
    \sum_{i \in V} x_{ij} \leq 1\ \ \forall\ j \in V 
\end{equation}

\textbf{Out-degree constraint}: Similar to the previous constraint this enforces that each node should be left from no more than one node. If the vehicle actually leaves to a node $j$, this will be $1$. Otherwise if the vehicle never visits $i$ or $j$, then this summation would be 0.
\begin{equation}
    \sum_{j \in V} x_{ij} \leq 1\ \ \forall\ i \in V 
\end{equation}

\textbf{Valid path constraint}:  This constraint ensures that the path chosen is valid. If it's a valid path the indegree at any vertex would be equal to the outdegree of that node. Otherwise, nodes would be acting as sources or sinks but this is possible only for the source or destination node.
\begin{equation}
    \sum_{i \in V} x_{ji} + w^S_i = \sum_{i \in V} x_{ij} + w^D_i\ \ \forall\ j \in V
\end{equation}

The next constraint defines $q_j$, the charge level when leaving station $j$ in terms of the charge level at station $i$, the depletion when traveling from $i$ to $j$, and the amount of charging done at $j$.

\textbf{State-of-Charge constraint}: This constraint sets up our charging levels according to our battery depletion equations. The charge level when the vehicle leaves each node $i$ is hence stored in $q_i$.
\begin{equation}
    (q_j - q_i) x_{ij}  = \bigg(z_j*y_j - \frac{(z_j \delta_j + d_{ij})}{C\gamma} \bigg) x_{ij} \ \forall\ i,j \in V
\end{equation}

\textbf{State-of-Charge constraint (source)}: Same as previous equation but defined separately for the source node.

\begin{equation}
    (q_i - \alpha) w^S_i  = \bigg(z_i*y_i - \frac{(z_i \delta_i + d_{Si})}{C\gamma} \bigg) w^S_i  \ \forall\ i \in V
\end{equation}

\textbf{State-of-Charge constraint (destination)}: Same as previous equation but defined separately for the destination.
\begin{equation}
    (\beta - q_i) w^D_i  = \bigg( - \frac{d_{iD}}{C\gamma} \bigg) w^D_i  \ \forall\ j \in V
\end{equation}

\textbf{Node reachability constraint}: This node actually ensures that the battery of the vehicle does not get depleted before it can reach a nearby station.

\begin{equation}
   \bigg(\frac{(z_j \delta_j + d_{ij})}{C\gamma}  - q_i\bigg) x_{ij} \leq \epsilon  \ \forall\ i,j \in V
\end{equation}
where $\epsilon$ is a very small number. \\

\textbf{Bound constraints}:
\begin{equation}
    x_{ii} = 0\ \ \forall\ i \in V
\end{equation}
\begin{equation}
    0 \leq y_i \leq 1\ \ \forall\ i \in V
\end{equation}
\begin{equation}
    0 \leq q_i \leq 1\ \ \forall\ i \in V
\end{equation}
\begin{equation}
    0 \leq \beta \leq 1
\end{equation}
\begin{equation}
    w^S_i \in \{0, 1\}\ \ \forall\ i \in V
\end{equation}
\begin{equation}
    w^D_i \in \{0, 1\}\ \ \forall\ i \in V
\end{equation}
\begin{equation}
    z_i \in \{0, 1\}\ \ \forall\ i \in V
\end{equation}
\begin{equation}
    x_{ij} \in \{0, 1\}\ \ \forall\ i,j \in V
\end{equation}

\subsection{Analysis}
\subsubsection{Constraints} \label{subsec:constraints}

\begin{enumerate}

    \item \textbf{In-degree constraint}: Each node has 1 such constraint, total of $|V|$ constraints.
    \item \textbf{Out-degree constraint}: Each node has 1 such constraint, total of $|V|$ constraints.
    \item \textbf{Valid path constraint}: Each node has 1 such constraint, total of $|V|$ constraints.
    \item \textbf{State-of-Charge constraint}: Each pair of nodes has 1 such constraint, total of $|V|^2$ constraints.
    \item \textbf{Node reachability constraint}: Each pair of nodes has 1 such constraint, total of $|V|^2$ constraints.
    
\end{enumerate}
Therefore, we end up with $2|V|^2 + 3|V|$ constraints in total, i.e., $\mathcal{O}(|V|^2)$. \\

\subsubsection{Variables}

\begin{enumerate}
    \item We define $|V|$ variables for the amount of charge added ($y$) at each node .
    \item $|V|$ variables for the amount of charge when leaving each node ($q$).
    \item We also define $|V|^2$ binary variables that indicate if the edge $(i, j)$ was traversed.
    \item $|V|$ variables for if a charging station was visited ($z_i$).
    \item  $2|V|$ variables for the in and out sum at each node.
    \item Finally, $2|V|$ more variables for the node taken from source and to destination respectively.
\end{enumerate}

Therefore, we have $|V|^2 + 7|V|$ variables that need to be decided, i.e., $\mathcal{O}(|V|^2)$.

\section{Dataset Generation}
\label{sec:dataset}
For testing our solution, we generate 4 datasets with varying levels of difficulty.

We first describe the parameters used in generating the dataset and the distribution from which they were sampled.

\begin{itemize}
    \item \textbf{Route graph}: We need to generate a Directed Acyclic Graph (DAG) that appropriately represents real-world road data. To do so, we generate the graph structure based on 3 parameters - number of levels, maximum nodes on a level, and probability of existence of edge. 
    \begin{enumerate}
        \item The first parameter defines the number of levels present in the DAG, the higher the value of this, the longer the route between source and destination. This has been varies between $[2, 4, 6, 8]$ to generate datasets of different difficulties.
        \item  The second parameter controls the number of possible routes that a vehicles can take. A higher value of this parameter increases the possible number of routes that the vehicles can take and also increases the number of parameters. For each level, the number of nodes on that level is picked uniformly from the range $[1, \text{Max Nodes}]$.
        \item The third parameter influences the density of the resulting DAG, a lower value implies that the probability of an edge existing between two nodes is lower. Hence, the graph would be sparser. 
    \end{enumerate}
    \item \textbf{Distance matrix} - The distance matrix has values sampled uniformly from a normal distribution with a mean of 300 km and a standard deviation of 50. Hence, $d_{ij} \in \mathcal{N}(300, 50)$. This is then multiplied element wise with the adjacency matrix of the graph, and the elements with 0 value are set to a very large value to indicate the non existence of an edge. The resulting matrix is the distance matrix of the graph
    \item \textbf{Distance from/to sources/sinks}: Similar to the distance matrix these values are also sampled uniformly from a Gaussian distribution with mean 300 and standard deviation of 50. 
    \item \textbf{Level of charging}: Charging stations are typically of 3 types - low(L1), medium(L2), and high (L3). These are chosen at uniformly at random.
    \item \textbf{Waiting times}: The waiting times at each station are also sampled from a normal distribution with mean 1 Hour and standard deviation 0.1 (6 minutes). These values were determined to be reasonable from our research.
    \item \textbf{Detour distances}: These distances are sampled from a Normal distribution with mean 10Km and standard deviation 1Km. 
    \item \textbf{Charging costs}: The rate of 1 unit of electricity in the US is around 0.134\$. Hence, we generate $|V|$ such values by sampling from a normal distribution with mean = 0.134\$ and standard deviation = 0.02. 
    
\end{itemize}

\begin{algorithm}[htbp]
 \caption{Route graph generation algorithm}
\small
 \begin{algorithmic}
\renewcommand{\algorithmicrequire}{\textbf{Input:}}
 \renewcommand{\algorithmicensure}{\textbf{Output:}}
 \Require Number of levels of DAG($n$), Max nodes per level ($D$)
 \Ensure  $DAG(\mathcal{V},\mathcal{E})$
    \State $N_{0}$ = $\{V_k\ for k \in \ randi(1, D)\}$
    \State $N\_prev \gets N_{0}$
     \State $\mathcal{V} = \mathcal{V} \cup N_0$
    \For{$level \in 1..n-2 $} 
        \State  $N_{level}$ = $\{V_k for k \in \ randi(1, D)\} $
        \State $edges \gets \{\}$
        \For{$i \in N\_prev $}
            \For{$j \in N_{level} $}    
                \If{$rand() \leq p_{edge}$}
                    \State $edges = edges \cup \{V_i, V_j\}$
                \EndIf
            \EndFor
        \EndFor
        \If{$edges = \phi$}
            \State Add a random edge to $edges$
        \EndIf
        \State $\mathcal{E} = \mathcal{E} \cup edges$
    \EndFor
\end{algorithmic}
\end{algorithm} 

\begin{figure}
    \centering
    \includegraphics[width = 0.5\textwidth]{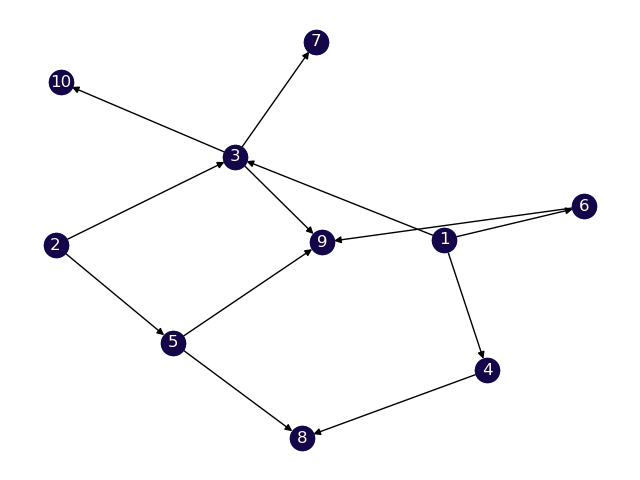}
    \caption{Instance 1 with 10 nodes and 2 levels.}
    \label{fig:inst1}
\end{figure}

\begin{figure}
    \centering
    \includegraphics[width = 0.5\textwidth]{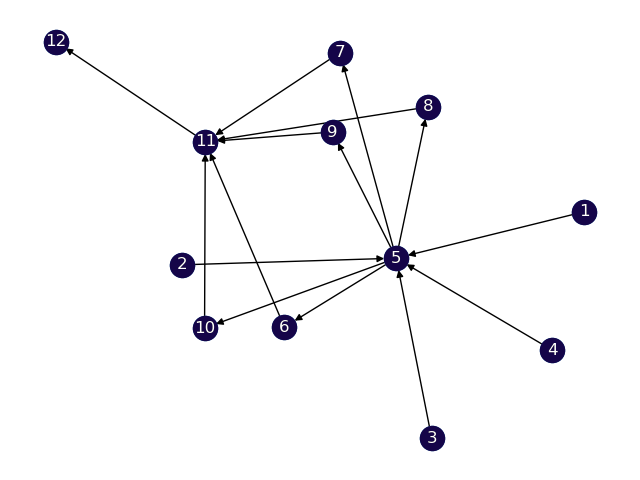}
    \caption{Instance 2 with 12 nodes and 4 levels.}
    \label{fig:inst2}
\end{figure}

\begin{figure}
    \centering
    \includegraphics[width = 0.5\textwidth]{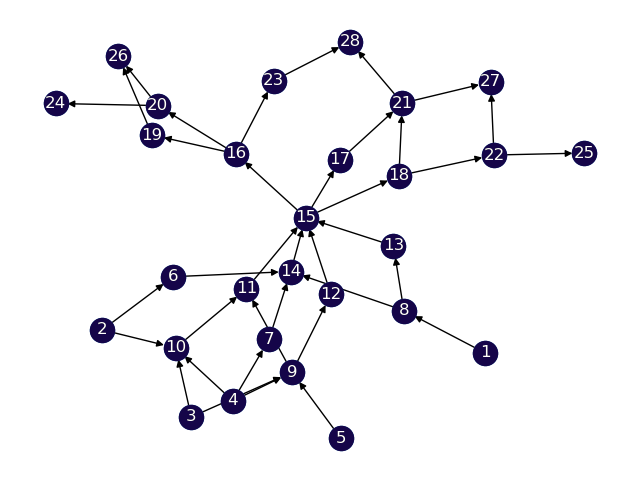}
    \caption{Instance 3 with 28 nodes and 6 levels.}
    \label{fig:inst3}
\end{figure}

\begin{figure}
    \centering
    \includegraphics[width = 0.5\textwidth]{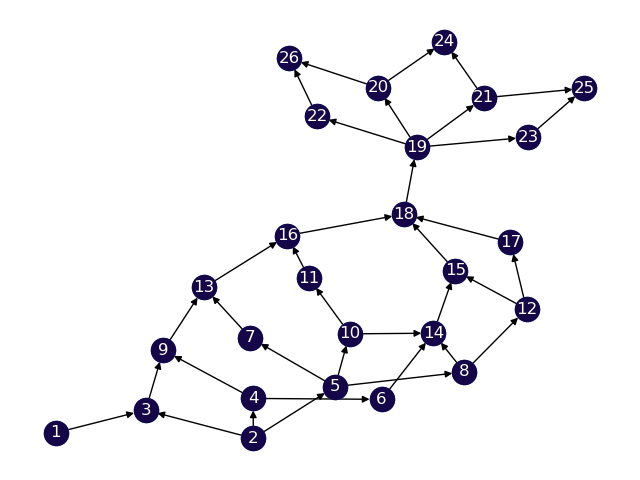}
    \caption{Instance 4 with 26 nodes and 8 levels.}
    \label{fig:inst4}
\end{figure}

The 4 instances used for testing our model are specified in Figure \ref{fig:inst1} - \ref{fig:inst4}. 


\section{Mathematical Programming}
\label{sec:mathprog}
The formulated problem is a multi-objective MINLP. But, before we solve the multi-objective problem, we solve two single objective optimization problems, one for minimizing time, and the other for minimizing cost. We feed each single objective function along with all the constraints in GAMS and use different solvers to arrive at a solution. 

We use the following solvers:
\begin{itemize}
    \item \textbf{ANTIGONE:} A deterministic global optimization for MINLP.
    \item \textbf{BARON:} Branch-And-Reduce Optimization Navigator for proven global solutions.
    \item \textbf{KNITRO:} Large scale NLP solver.
    \item \textbf{LINDOGLOBAL:} MINLP solver for proven global solutions.
    \item \textbf{OCTERACT:} MINLP solver for proven global solutions.
    \item \textbf{XPRESS:} High performance and fast LP/MIP and SLP based MINLP solver.
\end{itemize}

\begin{figure}[!ht]
    \centering
    \includegraphics[width = 0.5\textwidth]{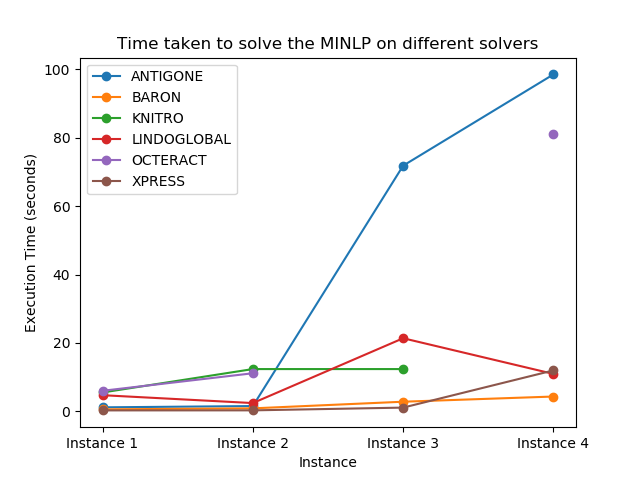}
    \caption{Comparison of execution time of various solvers for MINLP in GAMS.}
    \label{fig:time}
\end{figure}

We use the NEOS server to find solutions for each of the 4 test instances. The results for all the instances are summarized in Table \ref{tab:res}. In addition, the execution time for each of the solvers is compared in Figure \ref{fig:time}.
% Figure \ref{fig:obj_time} and Figure \ref{fig:obj_cost} represent the values of the two objective function achieved by different solvers across the 4 instances. 

% \setlength{\tabcolsep}{20pt}

\begin{table}[]
    \caption{Parameter Values}
    \centering
    \begin{tabular}{|c|c|}
    \hline
         \textbf{Parameter} & \textbf{Value} \\
         \hline
         Average speed of the vehicle & 50 (Km/hr)\\
         \hline
         Battery Capacity & 100 (KWh)\\
         \hline
         Mileage & 6 (Km/KWh) \\
         \hline
         Initial State-of-Charge & 1 \\
         \hline
    \end{tabular}
    \label{tab:parameters}
\end{table}

\begin{table*}[htbp]
     \caption{Objective function values from different solvers by individually optimizing both objectives}
    \centering
    \begin{tabular}{|c|c|c|c|c|c|c|c|c|}
    \hline
    \multirow{2}{*}{Solver} & \multicolumn{2}{|c|}{Instance 1}& \multicolumn{2}{|c|}{Instance 2}& \multicolumn{2}{|c|}{Instance 3} & \multicolumn{2}{|c|}{Instance 4}\\
    \cline{2-9}
    & Time & Cost & Time & Cost & Time & Cost & Time & Cost \\
    \hline
    ANTIGONE & 64.5472 & 80.1888 & 90.3843 & 176.8238 & 152.4195 & 255.598 & 196.059 & 335.1289 \\ \hline
    BARON & 64.5472 & 80.1888 & 90.3843 & 176.8238 & 152.4195 & 255.598 & 196.059 & 335.1289 \\ \hline
    KNITRO & 88.5455 & 98.9038 & 94.1911 & 206.8246 & $\sim$ & $\sim$ & $\sim$ & $\sim$ \\ \hline
    LINDOGLOBAL & 64.5472 & 80.1888 & 90.3843 & 176.8238 & 152.4195 & 271.0598 & 196.059 & 335.1289 \\ \hline
    OCTERACT & 64.5472 & 80.1888 & 90.3843 & 176.8238 & $\sim$ & $\sim$ & 196.059 & 335.1289 \\ \hline
    XPRESS & 98.7918 & 80.1888 & 103.5049 & 186.0169 & 169.2109 & 266.2965 & 196.059 & 335.1289 \\ 
\hline

    \end{tabular}
    \label{tab:res}
\end{table*}

% \begin{figure}[!ht]
%     \centering
%     \includegraphics[width = 0.5\textwidth]{images/obj_time.png}
%     \caption{Objective function values for time in single objective optimization using GAMS.}
%     \label{fig:obj_time}
% \end{figure}

% \begin{figure}[!ht]
%     \centering
%     \includegraphics[width = 0.5\textwidth]{images/obj_cost.png}
%     \caption{Objective function values for cost in single objective optimization using GAMS.}
%     \label{fig:obj_cost}
% \end{figure}

\subsection{Multi-Objective optimization}
We now extend our solutions of the single objective optimization problems to a multi-objective solution that minimizes both time and cost of charging. The MINLP solvers available on GAMS work for a single objective only, hence we use the $\epsilon$-constraint method to apply the single objective solver as done in \cite{mavrotas2009effective}. We provide the algorithm used for applying the $\epsilon$-constraint method in Algorithm \ref{alg:eps}.

\begin{algorithm}[htbp]
 \caption{$\epsilon$ - constraint method}
\label{alg:eps}
\small
 \begin{algorithmic}[1]
\renewcommand{\algorithmicrequire}{\textbf{Input:}}
 \renewcommand{\algorithmicensure}{\textbf{Output:}}
 \Require Fitness functions $f_1, f_2$, $\Delta$
 \Ensure  Pareto front  $\mathcal{P}$
 \State $x_1^{best} \gets Minimize(P, f_1)$
 \State $\mathcal{P} \gets \{x_2^{best}\}$
\State $\epsilon \gets f_2(x_1^{best}) + \Delta$
\While{$Minimize(P, f_1) \ni f_2(x) \leq \epsilon$}
\State $\hat{x} \gets Minimize(P, f_1) \ni f_2(x) \leq \epsilon$
\State $\mathcal{P} \gets \mathcal{P} \cup \hat{x}$
\State $\epsilon \gets f_2(x_1^{best}) + \Delta$
 \EndWhile 
 \State $\mathcal{P} \gets filter\_nondominated(\mathcal{P})$
    \end{algorithmic}
\end{algorithm}

\section{Results}
\label{sec:results}

\begin{figure}
    \centering
    \includegraphics[width = 0.5\textwidth]{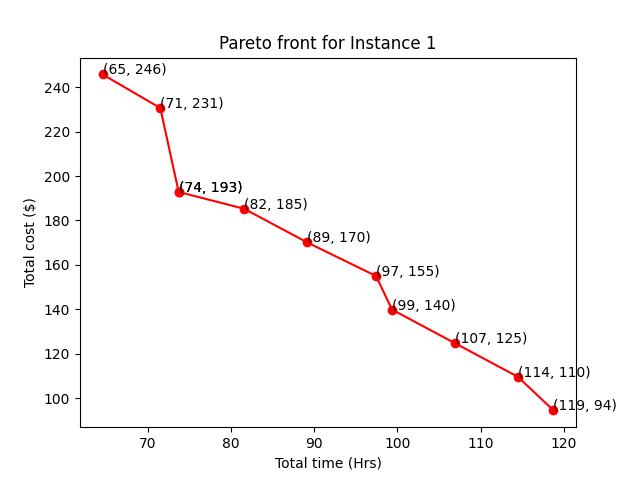}
    \caption{Pareto Front for Instance 1 obtained through OCTERACT solver in GAMS.}
    \label{fig:pareto1}
\end{figure}

\begin{figure}
    \centering
    \includegraphics[width = 0.5\textwidth]{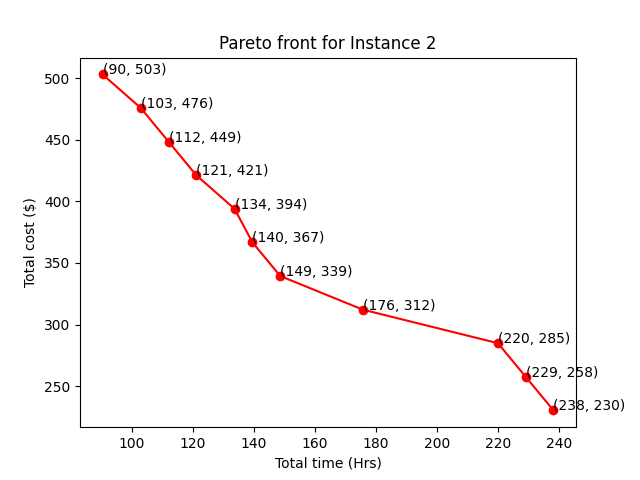}
    \caption{Pareto Front for Instance 2 obtained through OCTERACT solver in GAMS.}
    \label{fig:pareto2}
\end{figure}

\begin{figure}
    \centering
    \includegraphics[width = 0.5\textwidth]{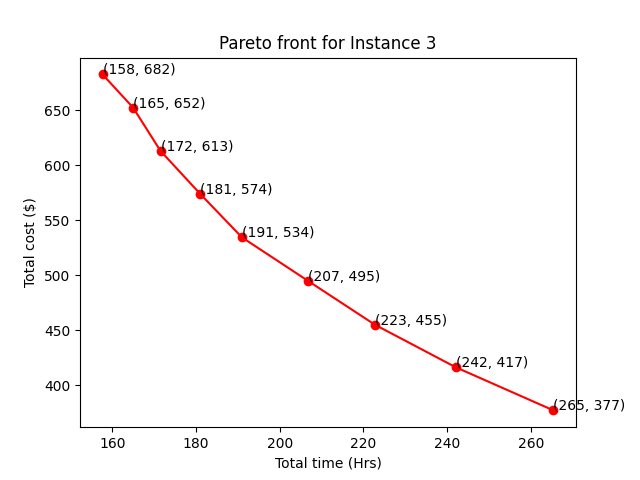}
    \caption{Pareto Front for Instance 3 obtained through LINDOGLOBAL solver in GAMS.}
    \label{fig:pareto3}
\end{figure}

\begin{figure}
    \centering
    \includegraphics[width = 0.5\textwidth]{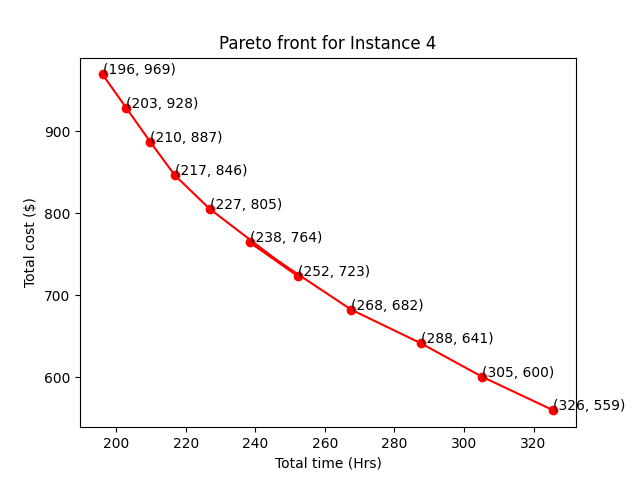}
    \caption{Pareto Front for Instance 4 obtained through LINDOGLOBAL solver in GAMS.}
    \label{fig:pareto4}
\end{figure}

On analyzing Table \ref{tab:res}\footnote{For all experiments, we use the parameter values as mentioned in Table \ref{tab:parameters}.}, we can infer that most solvers produce similar results for all instances. Since the values are solutions to single objective optimization problems, we can compare the solvers one-to-one for each objective function. The XPRESS and KNITRO solvers yield the least optimal route for the vehicle with respect to the total time taken to reach the destination. Although, from Figure \ref{fig:time}, it is evident that the XPRESS solver has the least execution time for most instances, alleging a trade-off between execution time and optimal objective function value. The ANTIGONE, BARON, and LINDOGLOBAL solvers effectively produce the most optimal results for all the data instances. Since, the BARON solver takes the least amount of time to execute among them, we conclude that it is the best solver for solving the formulated single objective MINLPs. We observe that as the instance size in terms of either total number of nodes or the number of levels increases, the time and cost of the journey increases, as expected. 

After solving single objective optimization problems, we obtain certain routes corresponding to each instance. But we extend our model to optimize multiple objectives, both time and cost. We apply the previously discussed $\epsilon$-constraint method and solve the multi-objective optimization problem using single-objective solvers. We implement the algorithm for our formulation and test it on all the data instances. We use the OCTERACT solver for instance 1 and instance 2, and the LINDOGLOBAL solver for instance 3 and instance 4. Due to the large size of the problem, we used the NEOS server for solving the MINLP. Figure \ref{fig:pareto1}, Figure \ref{fig:pareto2}, Figure \ref{fig:pareto3}, and Figure \ref{fig:pareto4} give the Pareto front obtained for data instances 1, 2, 3, and 4, respectively. 

From the shown Pareto fronts, it is fairly evident that there are no trivial solutions. For example, consider the points in the middle (89, 170) and (97, 155) in Figure \ref{fig:pareto1}. The user can easily make a choice between the two since $\$15$ is not worth spending 8 extra hours on the road. Similar explanations hold for other instances, and options on other Pareto fronts. 

Although we obtain these Pareto fronts, we are unsure about their global optimality since non linear constraints are considered. Hence, in search for an optimal solution to the problem formulation, we discuss meta heuristic techniques.

\section{Meta-heuristic Technique}
\label{sec:metaheuristic}
\begin{figure}[H]
    \centering
    \includegraphics[width=7cm]{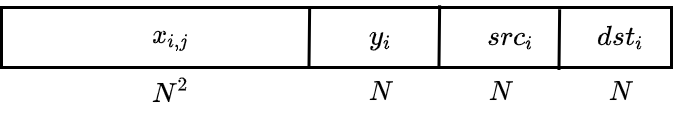}
    \caption{Solution structure}
    \label{fig:solution}
\end{figure}

We design an appropriate solution structure and use it in out meta-heuristic approach to represent the solutions. The algorithms are implemented using the \textit{mealpy} \cite{thieu_nguyen_2020_3711949} library in Python. Since, only continuous variables are allowed to be used in the algorithm, the binary decision variables used in the MP solution are reformulated in the form of floats in the range [0,1]. Appropriate methods for thresholding or other selection techniques to use them as binary decision variables have been employed. A solution structure has been displayed in Figure \ref{fig:solution}. 

The various binary variables have been manipulated to accommodate them as continuous variables. The selection technique is discussed below. 
\begin{enumerate}
    \item \textbf{Adjacency matrix} - The $x_{ij}$ values represent the adjacency matrix of the generated vehicle path. Ideally, we want only 0's and 1's in this matrix however due to library constraints these are continuous variables. The path is reconstructed from this float matrix using algorithm \ref{alg:routegenmeta}. The total number of variables contributed by this matrix is $|V|^2$.
    \item \textbf{Charge added} - The $y_{i}$ values indicate the amount of charge that was added in the $i^{th}$ station. Hence, there are $|V|$ such variables.
    \item \textbf{Source/Destination node} - These variables decide which node the vehicle takes from the source to enter the network, and which node it leaves through respectively. Combined, these contribute $2|V|$ variables.
\end{enumerate}

Hence, there are $|V|^2 + 3|V|$ variables in total. Once, these are generated and the path is decided we use the state-of-charge equations discussed in Section \ref{subsec:constraints} to update the battery state and objective function values at each point on the path. We adopt the penalty approach, and in case of any constraint violation we penalize the model with a large value (1e9). After the process the algorithm returns a valid path that does not violate any constraints (assuming that the final fitness is not very high).

\subsection{Genetic Algorithm}
The first type of meta-heuristic implemented is the classical Genetic Algorithm (GA) \cite{10.2307/24939139}. The algorithm initializes a random population, modifies the population using the crossover and mutation parameters. Finally, it adopts an elitist approach to pick best performing solutions from parents and offspring through a tournament. 
The crossover and mutation probability are both set to be $0.4$, also the number of epochs and population size are set to be $1000$.

\begin{figure}[!ht]
    \centering
    \includegraphics[width = 0.5\textwidth]{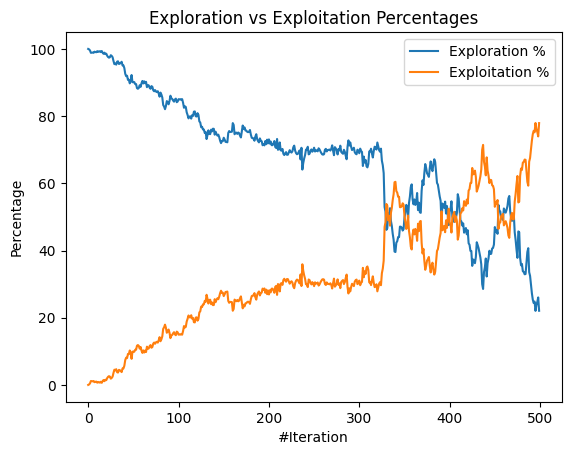}
    \caption{Exploration vs Exploitation curve for GA}
    \label{fig:eec_ga}
\end{figure}

\subsection{Particle Swarm Optimization}
The particle swarm based meta-heuristic, proposed by Eberhart and Kennedy \cite{494215} is one of the most popular computational intelligence techniques. It allows finer control on the solution exploration phase through the use of certain model parameters.
The search is governed by three parameters: $w$, $c_1$, $c_2$. Parameter $w$ controls the level of `inertia', a higher value leads to a more exploratory model. The $c_1$ and $c_2$ parameter govern the exploitation of the search space, $c_1$ nudges the model towards its locally best solution and $c_2$ nudges solution towards the globally best solution found by the swarm. Our solution linearly varies the $w$ value between $0.1$ and $0.5$ as the iterations progress. The $c_1$ and $c_2$ values are set to be $2.5$ and the number of epochs and population size are both set to $1000$.

\begin{figure}[!ht]
    \centering
    \includegraphics[width = 0.5\textwidth]{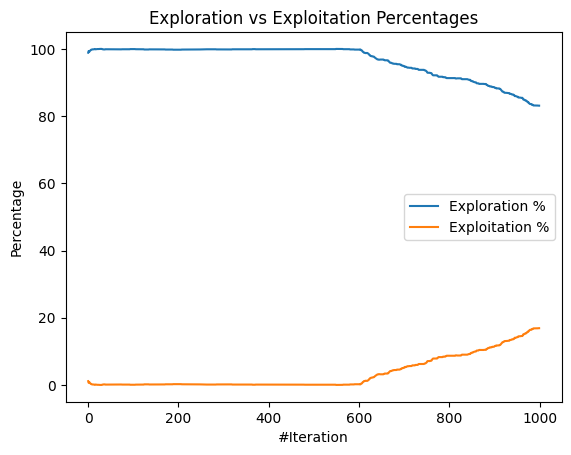}
    \caption{Exploration vs Exploitation curve for PSO}
    \label{fig:eec_pso}
\end{figure}

\subsection{Observations and Results}

\begin{figure}[!ht]
    \centering
    \includegraphics[width = 0.5\textwidth]{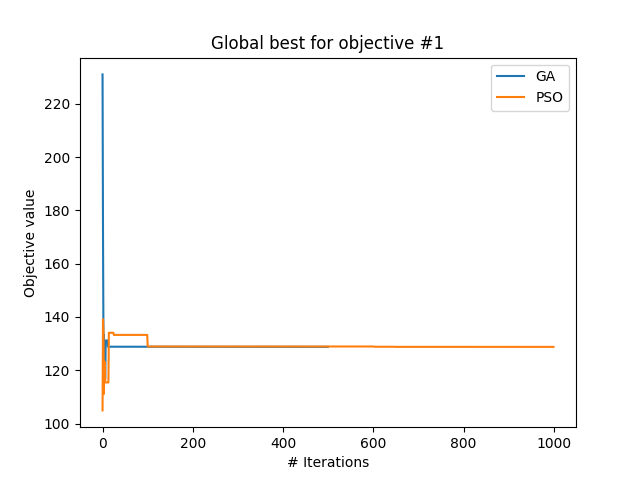}
    \caption{Convergence curve for time minimization}
    \label{fig:global obj 1}
\end{figure}

\begin{figure}[!ht]
    \centering
    \includegraphics[width = 0.5\textwidth]{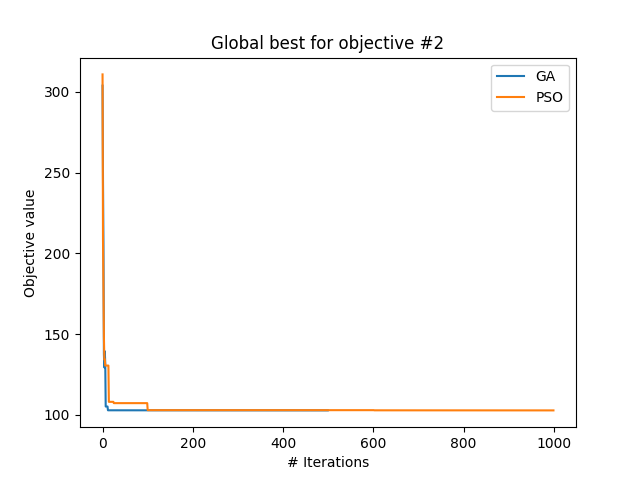}
    \caption{Convergence curve for cost minimization}
    \label{fig:global obj 2}
\end{figure}

We use Algorithm \ref{alg:mhroute} to generate the routes while using meta-heuristic approaches. On running both the techniques on data instance 1, we obtain various plots, and objective function values. The mealpy implementation in Python adds the two objective functions into one with equal weights and reports the solution corresponding to the minimum sum value. We run 2 independent runs. Figure \ref{fig:eec_ga} and Figure \ref{fig:eec_pso} explain the diversity of the solution over iterations for Run1. We observe that as the iterations progress, the exploration decreases while the exploitation increases. Figure \ref{fig:global obj 1} represents the convergence curve for the first objective function of total trip time, and Figure \ref{fig:global obj 2} represent the convergence curve for the second objective function of the cumulative cost of recharging, both for Run1. The objective function value obtained from the PSO meta-heuristic technique is [128.7, 102.7] and GA algorithm is [127.9, 103.1]. Another run of the implementation yielded [72.8, 192.7] for PSO, and [103.9, 165.6] for GA. The values obtained in Run1 are non-dominated with the Pareto points obtained for data instance 1 using the $\epsilon$-constraint method in Figure \ref{fig:pareto1}. We observe that the values obtained through PSO of Run2 is a dominating point of the Pareto front. Hence, we conclude that the meta-heuristic technique produces at least as good a solution as the $\epsilon$-constraint method. 

\begin{algorithm}[htbp]
 \caption{Meta-heuristic route generation}
\label{alg:mhroute}
\small
 \begin{algorithmic} \label{alg:routegenmeta}
\renewcommand{\algorithmicrequire}{\textbf{Input:}}
 \renewcommand{\algorithmicensure}{\textbf{Output:}}
 \Require Adjacency matrix ($X$)
 \Ensure  Path $\mathcal{P}$

     \State $src \gets argmax(src)$
    \State $ \mathcal{P} \gets \{src\}$
     \State $dest \gets argmax(dest)$
     \State $curr \gets src$
    
     \While{$curr \not = dest$}
        \State $next \gets argmax(X[curr])$
        \State $\mathcal{P} \gets  \mathcal{P} \cup {next}$
        \State $curr \gets next$
    \EndWhile
\end{algorithmic}
\end{algorithm}

\section{Conclusion}
\label{sec:conclusions}
The EVRP is a fairly new, though well-researched problem with a lot of variants in the form of varying objective functions, constraints, charging models, and pricing methods. We propose a problem formulation for personal EVRP (pEVRP) where the goal is to optimize both the time taken for a long origin destination trip and the cost of recharging on the way for a single EV vehicle. We incorporate real-life factors such as traffic delays, detour distances, and variable pricing to our problem. We formulate an MINLP for the same and define various constraints to ensure proper traversal of the underlying network graph. We extend the solutions of single objective MINLPs to multi-objective solutions by using the $\epsilon$-constraint method. We obtain a set of non-dominated points, i.e., obtain the first Pareto front. The results of this method are presented as figures portraying the obtained Pareto front. The solutions are non-trivial and the user can make a choice based on personal flexibility. We further apply two meta-heuristic techniques (GA and PSO) to obtain globally optimal solutions. We define a solution structure, and use Python to implement the meta-heuristic algorithms. Finally, we observe that the results of the $\epsilon$-constraint method are similar to those obtained from the meta-heuristic techniques. In some cases, the meta-heuristic technique works slightly better than the $\epsilon$-constraint method in finding more optimal solutions. 

% \section{Future Work}
% \label{sec:future}
% One of the central points of improvement could be the use of a more refined version of meta-heuristic techniques. GA and PSO can be tweaked and experimented with to obtain more optimal solutions. We can also add a battery health optimization function due to which the optimal charging stops as well as the amount of charge added at the stop would ensure healthy battery life. We can explore the possibility of adding non-linear charging of EVs. 

\bibliographystyle{IEEEtran}
\bibliography{main}

\begin{thebibliography}{10}
\providecommand{\url}[1]{#1}
\csname url@samestyle\endcsname
\providecommand{\newblock}{\relax}
\providecommand{\bibinfo}[2]{#2}
\providecommand{\BIBentrySTDinterwordspacing}{\spaceskip=0pt\relax}
\providecommand{\BIBentryALTinterwordstretchfactor}{4}
\providecommand{\BIBentryALTinterwordspacing}{\spaceskip=\fontdimen2\font plus
\BIBentryALTinterwordstretchfactor\fontdimen3\font minus
  \fontdimen4\font\relax}
\providecommand{\BIBforeignlanguage}[2]{{%
\expandafter\ifx\csname l@#1\endcsname\relax
\typeout{** WARNING: IEEEtran.bst: No hyphenation pattern has been}%
\typeout{** loaded for the language `#1'. Using the pattern for}%
\typeout{** the default language instead.}%
\else
\language=\csname l@#1\endcsname
\fi
#2}}
\providecommand{\BIBdecl}{\relax}
\BIBdecl

\bibitem{LIN2016508}
\BIBentryALTinterwordspacing
J.~Lin, W.~Zhou, and O.~Wolfson, ``Electric vehicle routing problem,''
  \emph{Transportation Research Procedia}, vol.~12, pp. 508--521, 2016, tenth
  International Conference on City Logistics 17-19 June 2015, Tenerife, Spain.
  [Online]. Available:
  \url{https://www.sciencedirect.com/science/article/pii/S2352146516000089}
\BIBentrySTDinterwordspacing

\bibitem{felipe2014heuristic}
{\'A}.~Felipe, M.~T. Ortu{\~n}o, G.~Righini, and G.~Tirado, ``A heuristic
  approach for the green vehicle routing problem with multiple technologies and
  partial recharges,'' \emph{Transportation Research Part E: Logistics and
  Transportation Review}, vol.~71, pp. 111--128, 2014.

\bibitem{paz2018multi}
J.~Paz, M.~Granada-Echeverri, and J.~Escobar, ``The multi-depot electric
  vehicle location routing problem with time windows,'' \emph{International
  journal of industrial engineering computations}, vol.~9, no.~1, pp. 123--136,
  2018.

\bibitem{KUCUKOGLU2021107650}
\BIBentryALTinterwordspacing
I.~Kucukoglu, R.~Dewil, and D.~Cattrysse, ``The electric vehicle routing
  problem and its variations: A literature review,'' \emph{Computers \&
  Industrial Engineering}, vol. 161, p. 107650, 2021. [Online]. Available:
  \url{https://www.sciencedirect.com/science/article/pii/S0360835221005544}
\BIBentrySTDinterwordspacing

\bibitem{ERDOGAN2012100}
\BIBentryALTinterwordspacing
S.~Erdoğan and E.~Miller-Hooks, ``A green vehicle routing problem,''
  \emph{Transportation Research Part E: Logistics and Transportation Review},
  vol.~48, no.~1, pp. 100--114, 2012, select Papers from the 19th International
  Symposium on Transportation and Traffic Theory. [Online]. Available:
  \url{https://www.sciencedirect.com/science/article/pii/S1366554511001062}
\BIBentrySTDinterwordspacing

\bibitem{ding2015conflict}
N.~Ding, R.~Batta, C.~Kwon \emph{et~al.}, ``Conflict-free electric vehicle
  routing problem with capacitated charging stations and partial recharge,''
  \emph{SUNY, Buffalo}, 2015.

\bibitem{kancharla2020electric}
S.~R. Kancharla and G.~Ramadurai, ``Electric vehicle routing problem with
  non-linear charging and load-dependent discharging,'' \emph{Expert Systems
  with Applications}, vol. 160, p. 113714, 2020.

\bibitem{liu2014joint}
C.~Liu, J.~Wu, and C.~Long, ``Joint charging and routing optimization for
  electric vehicle navigation systems,'' \emph{IFAC Proceedings Volumes},
  vol.~47, no.~3, pp. 2106--2111, 2014.

\bibitem{sachenbacher2011efficient}
M.~Sachenbacher, M.~Leucker, A.~Artmeier, and J.~Haselmayr, ``Efficient
  energy-optimal routing for electric vehicles,'' in \emph{Twenty-fifth AAAI
  conference on artificial intelligence}, 2011.

\bibitem{de2013intention}
M.~M. De~Weerdt, E.~Gerding, S.~Stein, V.~Robu, and N.~R. Jennings,
  ``Intention-aware routing to minimise delays at electric vehicle charging
  stations,'' 2013.

\bibitem{mavrotas2009effective}
G.~Mavrotas, ``Effective implementation of the $\varepsilon$-constraint method
  in multi-objective mathematical programming problems,'' \emph{Applied
  mathematics and computation}, vol. 213, no.~2, pp. 455--465, 2009.

\bibitem{thieu_nguyen_2020_3711949}
\BIBentryALTinterwordspacing
N.~V. Thieu, ``A collection of the state-of-the-art meta-heuristic algorithms
  in python: Mealpy,'' 2020. [Online]. Available:
  \url{https://doi.org/10.5281/zenodo.3711948}
\BIBentrySTDinterwordspacing

\bibitem{10.2307/24939139}
\BIBentryALTinterwordspacing
J.~H. Holland, ``Genetic algorithms,'' \emph{Scientific American}, vol. 267,
  no.~1, pp. 66--73, 1992. [Online]. Available:
  \url{http://www.jstor.org/stable/24939139}
\BIBentrySTDinterwordspacing

\bibitem{494215}
R.~Eberhart and J.~Kennedy, ``A new optimizer using particle swarm theory,'' in
  \emph{MHS'95. Proceedings of the Sixth International Symposium on Micro
  Machine and Human Science}, 1995, pp. 39--43.

\end{thebibliography}

\end{document}